\documentclass[11pt,a4paper]{article}
\usepackage[hyperref]{clib_acl}
\setcitestyle{notesep={: }}

\usepackage{times}
\usepackage{latexsym}
\usepackage{graphics,graphicx}

\usepackage[title]{appendix}
\usepackage{amsfonts}

\usepackage{microtype}

\aclfinalcopy 

\title{An Open-Domain QA System for e-Governance}

\author{Radu Ion, Andrei-Marius Avram, Vasile Păiș, Maria Mitrofan, \\
  \textbf{Verginica Barbu Mititelu, Elena Irimia and Valentin Badea} \\
  Research Institute for AI ``Mihai Drăgănescu'' \\
  13 ``Calea 13 Septembrie'' \\
  Bucharest 050711, Romania \\
  \texttt{\{radu,andrei.avram,vasile,maria\}@racai.ro} \\
  \texttt{\{vergi,elena,valentin.badea\}@racai.ro}}

\date{}

\begin{document}
\maketitle
\begin{abstract}
The paper presents an open-domain Question Answering system for Romanian, answering COVID-19 related questions. The QA system pipeline involves automatic question processing, automatic query generation, web searching for the top $10$ most relevant documents and answer extraction using a fine-tuned BERT model for Extractive QA, trained on a COVID-19 data set that we have manually created. The paper will present the QA system and its integration with the Romanian language technologies portal RELATE, the COVID-19 data set and different evaluations of the QA performance.

\end{abstract}

\section{Introduction}
\label{sec:intro}

According to \citet[1]{fengbin2021qasurvey}, open-domain Question Answering (QA) has the ability ``to answer a given question without any specified context'', by searching for the relevant documents on the web and extracting the relevant answer from one (or more) of the retrieved documents. In contrast, Machine Reading Comprehension ``aims to enable machines to read and comprehend specified context passage(s) for answering a given question'' which entails that, given a question and one (or more) passage(s) of text that (can) contain the answer, the QA system is able to identify it in the given text piece. The ``open-domain'' designation of a QA system also pertains to the ability of the system to answer factoid questions (factual questions) from any domain, according to \citet[1]{lewis20qatraintest}.

The QA system that is presented in this paper is ``open-domain'' from both points of view: it only takes the input question and automatically searches for the relevant documents on the web but, for the answer selection, it employs a fine-tuned BERT model for Extractive QA that, using the input question together with the snippet that the web search engine produces for each relevant document, highlights the answer to the input question. Although we present an instance of this system for the COVID-19 domain, given other targeted data sets, the exact same pipeline can be applied to answer questions from those domains.

The QA system was developed in the European project Enrich4All\footnote{\url{https://www.enrich4all.eu/}}, a project aiming at a Digital Single Market strategy, which is linked with lowering language barriers for online services and public administration procedures. The architecture of the QA system enables it to answer administrative questions about a certain topic (e.g. COVID-19, construction permits, etc.) that citizens may have for public authorities, by automatically searching for relevant documents on the public authority web site. Being available 24/7, it has the potential to reduce the administrative burden for public authorities.

In what follows, we present related approaches to open-domain QA in Section \ref{sec:related}, followed by a description of the COVID-19 data set in Section \ref{sec:dataset}. Section \ref{sec:bert} details the fine-tuning of different Romanian BERT models to COVID-19 Extractive QA, while Section \ref{sec:pipline} describes the architecture and the underpinnings of the open-domain QA system. We end the paper with Section \ref{sec:eval} devoted to the evaluation of the QA system and Section \ref{sec:conclusions} presenting concluding remarks and future work plans.

\section{Related work}
\label{sec:related}

Open-domain QA (ODQA) aims at answering questions from large open-domain corpora (e.g., Wikipedia). Recent success in this field mainly comes from fine-tuning and improving the pre-trained LMs, like ELMo \cite{peters1802deep} and BERT \cite{devlin2018bert}. 

\newcite{wang2019multi} proposed a multi-passage BERT model to globally normalize answer scores across all passages of the same question, enabling the QA model to find more precise answers utilizing more text passages. Splitting articles into passages with the length of 100 words improved performance by 4\% and trained on the OpenSQuAD data set, the model gained 21.4\% EM and 21.5\% F1 over all non-BERT models, and 5.8\% EM (exact match) and 6.5\% F1 over BERT-based models. 

\newcite{yang2019end} integrated BERT with the open-source Anserini information retrieval toolkit. They showed that combining a BERT-based reader with passage retrieval using the Anserini IR toolkit yields towards an improvement in question answering directly from a Wikipedia corpus. During training, passages corresponding to the same question are taken as independent training instances. The authors report that fine-tuning pre-trained BERT with SQuAD is sufficient to achieve high accuracy in identifying answer spans.

\newcite{lee2019latent} showed that it is sub-optimal to incorporate a standalone IR system in an OpenQA system, therefore they developed and they develop an OpenRetrieval Question Answering system (ORQA) system that treats the document retrieval from the information source as a latent variable and trains the whole system only from question-answer string pairs based on BERT. The system was evaluated on open versions of five QA data sets and  outperformed BM25 model by up to 19 points in exact match.

\newcite{karpukhin2020dense} used BERT pre-trained model \cite{devlin2018bert} and a dual-encoder architecture \cite{bromley1993signature} in order to develop a training scheme that uses a relatively small number of question and passage pairs. The authors demonstrated that by fine-tuning the question and passage encoders on existing question-passage pairs the system outperformed models, such as TF-IDF or BM25 and also that applying a reader model to the retrieved passages leads to comparable or better results on multiple QA data sets in the open-retrieval setting. Furthermore, the study showed that in the context of open-domain question answering, a higher retrieval precision translates to a higher end-to-end QA accuracy. 

\newcite{guu2020realm} used contextualized word representations to predict a span as answer. The authors showed the effectiveness of Retrieval-Augmented Language Model pre-training (REALM) by fine-tuning on the task of ODQA. The system outperformed previous methods by a significant margin (4-16\% absolute accuracy),and also provided qualitative benefits such as interpretability and modularity. 

\newcite{yamada2021efficient} introduced  Binary Passage Retriever (BPR), a memory-efficient neural retrieval model that integrates a learning-to-hash technique into a Dense Passage Retriever (DPR) \cite{karpukhin2020dense}. BPR has two main objectives: to generate efficient candidates based on binary codes and re-ranking based on continuous vectors. When compared with DPR, BPR reduced the memory cost from 65GB to 2GB without a loss of accuracy.

\section{The COVID-19 data set}
\label{sec:dataset}

The COVID-19 data sets we designed are a small corpus and a question-answer data set. The targeted sources were official websites of Romanian institutions involved in managing the COVID-19 pandemic, like The Ministry of Health, Bucharest Public Health Directorate, The National Information Platform on Vaccination against COVID-19, The Ministry of Foreign Affairs, as well as of the European Union. We also harvested the website of a non-profit organization initiative, in partnership with the Romanian Government through the Romanian Digitization Authority, that developed an ample platform with different sections dedicated to COVID-19 official news and recommendations. News websites were avoided due to the volatile character of the continuously changing pandemic situation, but a reliable source of information was the website of a major private medical clinic
, that provided detailed medical articles on important subjects of immediate interest for the readers and patients, like immunity, the emergent treating protocols, or the new 
variants of the virus.

Both the corpus and the question-answer data set were manually collected and revised. Data was checked for grammatical correctness and missing diacritics were introduced. The corpus is structured in 55 UTF-8 documents and contains 147,297 words.

The question-answer data set comprises 185 entries made up of a label (see the list of labels below), a question and its answer. 
The questions have been multiplied manually: rephrasing techniques have been used (such as active-passive constructions, personal-impersonal ones, constructions with or without (epistemic and/or deontic) modality, synonyms, antonyms, hypernyms, etc.), always making sure the question’s meaning is not altered, so that the existing answer could be appropriate for each of the resulted questions. All these diverse ways of expressing the same question are meant to serve as a train base for the BERT model so that it can recognize alternative ways of inquiring about a certain topic.

Each entry in the question-answer data set was associated with one of the labels: \texttt{covid-spread}, \texttt{covid-symptoms}, \texttt{covid-treatment}, \texttt{covid-vaccination}, \texttt{covid-logistics}, \texttt{covid-passport}, \texttt{covid-testing} and \texttt{covid-others}. 

To use the BERT model as a QA system (see Section \ref{sec:bert}), we had to manually mark the relevant answer to the question in the provided answer paragraph. An example of such an entry in the question-answer data set is given below: three different formulations of the same question, all marked with ``Q:'', synonyms for a word or phrase enumerated within the question, using square brackets and slashes, and finally, the enlarged answer comes last, marked with ``A:'', in which the more to-the-point answer is marked using square brackets as well:

\noindent \textbf{L}: \texttt{covid-spread}

\noindent \textbf{Q}: Vremea caldă \textbf{[previne/ne ferește de/ne protejează de]} infectarea cu Coronavirus? (``Does the warm weather \textbf{[prevent the/keep us safe from the/protect us against]} infection with the Coronavirus?'')

\noindent \textbf{Q}: Vara putem să \textbf{[facem/ne îmbolnăvim de]} COVID-19? (``Can we \textbf{[catch/get sick with]} COVID-19 in the summer?'')

\noindent \textbf{Q}: Dispare covidul \textbf{[pe vreme caldă/vara/la soare/la temperaturi mari]}? (``Does Covid vanish \textbf{[in warm weather/in the summer/in the sun/at high temperatures]}?'')

\noindent \textbf{A}: Datele existente arată că \textbf{[infecția poate fi dobândită în toate zonele climatice, inclusiv în cele calde]}. (``Existing data shows that \textbf{[the infection can be acquired in all climates, including the warm ones]}.'')

Starting from the example above we can generate $3 \cdot 2 \cdot 4 = 24$ different, but semantically equivalent formulations of the question ``Does the warm weather protect us from the Coronavirus?'' to which the answer is highlighted in the answer paragraph: ``Existing data shows that \textbf{[the infection can be acquired in all climates, including the warm ones]}.'' Having question-answer data points annotated in this way, we were able to automatically generate a SQuAD 2.0 data set \citep{rajpurkar2018squad} on which we fine-tuned a Romanian BERT model for Extractive QA, as described in the next section.

The COVID-19 SQuAD 2.0 data set\footnote{\url{https://github.com/racai-ai/e4a-covid-qa/tree/master/data}} contains 1,388 question-answer data points, after automatically expanding all possible question formulations as described above. Since each question entry has multiple formulations, in order to be fair to the Extractive learning model, we randomly set aside 10\% of a question alternative formulations for the test set (if there were less than 10 alternative formulations for a question, we kept a single formulation for the test set). This selection procedure gave us 180 question-answer data points in the development set and 1,208 question-answer data points in the training set, which represents a 13\%-87\% split of the data.

\section{Fine-tuning BERT for COVID-19 Extractive QA}
\label{sec:bert}

To create the QA model, we employed the standard BERT fine-tuning procedure described in \cite{kenton2019bert} that consists of putting two feed-forward layers on top of the contextualized embeddings to predict the start and the end of an answer. On a more granular level, this operation is equivalent to taking the dot product between either a start vector $S$ or an end vector $E$ and each of the contextualized embedding $T_i$ produced by the BERT model, and then applying the softmax function over the results:

\begin{equation}
    P(start_i) = \frac{e^{S \cdot T_i}}{\sum_{j}e^{S \cdot T_j}}
\end{equation}

\begin{equation}
    P(end_i) = \frac{e^{E \cdot T_i}}{\sum_{j}e^{E \cdot T_j}} 
\end{equation}

\noindent where $i$ is the index of the contextualized embedding. Then we select as answer to a question the span from $i$ to $j$ that maximizes the $S \cdot T_i + E \cdot T_j$, and that satisfies $j > i$ and $j - i < \xi$, where $\xi \in \mathbb{N}$ is a tunable hyperparameter that controls the maximum number of tokens admitted in a span.

\begin{table}[]
    \centering
    \begin{tabular}{lcc}
        \hline
        \textbf{Model} & \textbf{Exact \%} & \textbf{F1 \%} \\
        \hline
        BERT-base-ro-uncased & 71.33 & \textbf{77.25} \\
        BERT-base-ro-cased & \textbf{73.33} & 76.75 \\
        RoBERT-small & 58.00 & 61.64 \\
        RoBERT-medium & 59.33 & 63.06 \\
        RoBERT-large & 61.00 & 64.12 \\
        Distil-BERT-base-ro & 51.33 & 70.39 \\
        Distil-RoBERT-base & 55.33 & 61.72 \\
        DistilMulti-BERT-base-ro & 51.33 & 70.39 \\
        \hline
    \end{tabular}
    \caption{Results of the Romanian BERT models on our QA task}
    \label{tab:qa_results}
\end{table}

We fine-tuned eight Romanian BERT models on the question-answer data set introduced previously:

\begin{itemize}
    \item \textbf{BERT-base-ro-uncased} and \textbf{BERT-base-ro-cased}: the cased and uncased versions of the first Romanian BERT \cite{dumitrescu2020birth}.
    \item \textbf{RoBERT-small}, \textbf{RoBERT-medium} and \textbf{RoBERT-large}: the second iteration of Romanian BERTs introduced in \cite{masala2020robert}. In comparison with the first BERT models, the authors introduce a large and a small version, trained on different corpora.
    \item \textbf{Distil-BERT-base-ro}, \textbf{Distil-BERT-base-ro} and \textbf{DistilMulti-BERT-base-ro}: distilled version of Romanian BERTs \cite{avram2021distilling}. The first variant was obtained by distilling the knowledge of BERT-base-ro-cased, the second of RoBERT-base and the last of both BERT-base-ro-cased and RoBERT-base.
\end{itemize}

We trained each model for 5 epochs, using a batch size of 8, a learning rate of $3 \cdot 10^{-5}$, a weight decay of $10^{-3}$ and a maximum span $\xi = 30$ of tokens. We used the train-test split of the data set that was described in Section \ref{sec:dataset}. 

The results are depicted in Table \ref{tab:qa_results} where we outline the capabilities of the tested models to find the exact answer (i.e. Exact \%), and the overlap percentage between the predicted and the true spans (i.e. F1 \%). The highest exact score of 73.33\% was obtained by BERT-base-ro-cased and the highest F1-score by BERT-base-ro-uncased with 77.25\%. As it can be observed, there is a significant difference in performance between the RoBERT and BERT-base-ro variants. While the size of the test data is small (180 question-answer pairs), which could emphasize a disproportionate difference due to selection bias, another, more likely explanation could hold: the BERT-base-ro vocabulary contains 50K word pieces while RoBERT vocabulary contains 38K word pieces, which puts BERT-base-ro in a better position to cover COVID-19 vocabulary, thus making the fine-tuning process more successful.

\section{The open-domain QA pipeline}
\label{sec:pipline}

\begin{figure*}
    \centering
    \includegraphics[width=1.0\textwidth]{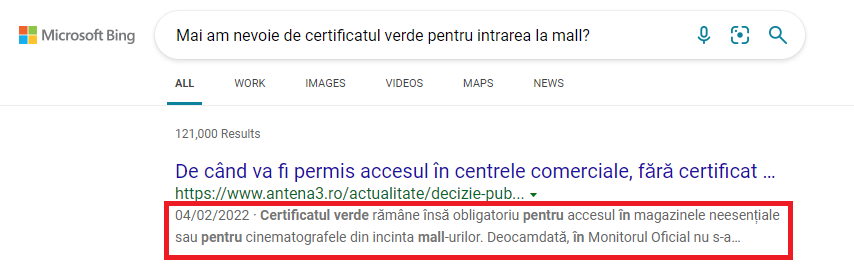}
    \caption{Bing search results}
    \label{fig:bingsnippet}
\end{figure*}

The trainable, open-domain QA system\footnote{\url{https://github.com/racai-ai/e4a-covid-qa}} executes the following operations, in sequence, for an input question:

\begin{itemize}
\item \textbf{Question processing}: the question string is run through the TEPROLIN Romanian text processing web service\footnote{\url{https://relate.racai.ro/index.php?path=teprolin/complete}} \citep{ion2018teprolin} to obtain tokenization, lemmatization and dependency parsing annotation.
\item \textbf{Query generation}: the question is analyzed to see which words are useful to form a query for the web search engine. Our web search engine of choice is Microsoft's Bing search engine\footnote{\url{https://www.microsoft.com/en-us/bing/apis/bing-web-search-api}} because it is one of the few that offers API-based querying and offers access to the search hits via a JSON object containing text snippets and URLs of the relevant documents. Subsection \ref{ssec:qgen} details how the query is formed from the processed version of the input question.
\item \textbf{Answer mining}: entailing web search results re-ranking and answer highlighting, for each hit of the Bing web search engine (out of the total 10 that we ask for), we call the previously described, fine-tuned QA BERT model with the input question and the Bing-found text snippet (see the red rectangle in Figure \ref{fig:bingsnippet}) as the question context and get back a highlight of the answer (within the snippet) that BERT model thinks is a right fit for the input question. The highlighting comes with a confidence measure, topping at 1 for certainty and going towards 0 when the confidence level drops. If the hit has rank $r$, $0 \leq r < 10$ provided by Bing and BERT's model confidence in highlighting the correct answer is $c$, $0 \leq c \leq 1$, then the combined confidence of the answer is
\begin{equation} \label{eq:3}
    q = c \cdot \frac{10-r}{10}
\end{equation}
\end{itemize}

\subsection{Query generation}
\label{ssec:qgen}

The query generation algorithm takes the processed input question and produces a list of query terms for the Bing search engine. The question has been processed with the TEPROLIN text processing web service and we have, thus, lemmas, parts of speech and dependency information for each token in the input question.

We have experimented with three query generation procedures:

\begin{itemize}
    \item \textbf{The baseline algorithm}: just take the input question as it is and feed it to the Bing search engine.
    \item \textbf{The content word selection algorithm}: take all nouns, numerals, verbs, adjectives and adverbs, in the order they appear in the question and form a maximally matching disjunctive query, e.g. for three terms $t_1, t_2, t_3$ the query is ``$t_1\ t_2\ t_3$''. Very frequent Romanian verbs are not included in the query, such as ``a avea'' (to have), ``a fi'' (to be), ``a exista'' (to exist), ``a face'' (to do), etc.
    \item \textbf{No diacritics content word selection algorithm}: the content word selection algorithm query from which we automatically remove the Romanian diacritics. We see that Romanian pages are written with or (more likely) without the proper diacritics and thus, we have to accommodate query terms with and without diacritics.
\end{itemize}

While Bing works surprisingly well with the baseline query algorithm (the input question), by empirical experimentation we find that results from the union of the content word selection algorithm and its ``no diacritics'' version provide a better ranking for the most relevant documents. For instance, for the question from Figure \ref{fig:bingsnippet} (``Do I need the COVID certificate to be allowed in the mall?''), the generated query is ``nevoie certificatul verde intrarea mall'' which produces 2 relevant documents on the first page, while the default query with just the input question yields a single relevant document on the first page. Furthermore, Bing does not filter out Romanian functional words, e.g. ``pentru'' (for), and considers them to be relevant (easily seen because these are in bold in the returned snippets).

\subsection{Integration with RELATE}
\label{ssec:relate}
RELATE \citep{pais-etal-2020-processing,pais2020multiplepipelinesrelate} is a modular platform allowing access through a web based interface to multiple natural language processing applications for Romanian language. It follows, in a simpler way, the European Language Grid\footnote{\url{https://www.european-language-grid.eu/}} philosophy of integrating components based on a micro-services architecture. In this context, the developed QA system was first exposed as a JSON REST API. This allowed it to be easily integrated in the RELATE platform and thus it became accessible through the platform's web front-end.

\begin{figure*}[th!]
    \centering
    \includegraphics[width=1.0\textwidth]{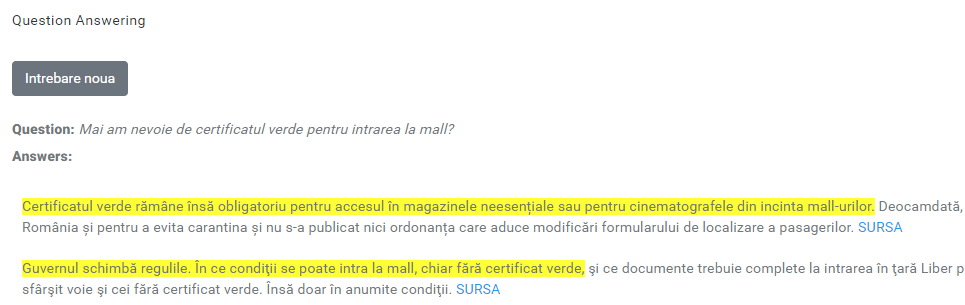}
    \caption{Question answers}
    \label{fig:relate_1}
\end{figure*}

The QA interface\footnote{\url{https://relate.racai.ro/index.php?path=qa/demo}} allows the user to enter a question, select the desired model\footnote{Currently, a single model is available.} and then pass the input data to the system. Results are displayed in the form of text snippets extracted from various Internet sources. The actual answer is highlighted in the text snippet and the user is given the opportunity to access the source web site associated with the snippet. Finally, the user can return to the previous page in order to ask a new question. The output of the QA interface is presented in Figure \ref{fig:relate_1}.

\section{Evaluation}
\label{sec:eval}

To fairly evaluate our QA system, we have developed a new test set, containing 65 COVID-19 related questions, that were created to be different from the ones in the data set presented in Section \ref{sec:dataset}. Thus, each of the authors of this paper independently recorded the most 10 interesting questions, from a personal point of view, mimicking real usage of the system. The intent behind this decision was to evaluate our QA system in real-world scenario where users may ask all sorts of questions, using Romanian diacritics or not, about a very rapidly evolving subject such as COVID-19. We have incrementally developed our COVID-19 data set in a time frame of about 6 months, in which time some understudied or evolving aspects of COVID-19 (e.g. the duration of the vaccine-induced immunity, the different vaccines efficiencies or the number of days the infected persons are quarantined) have changed, as results of ongoing studies were published that shed new light on these aspects.

We have found out that, for most of the questions, Bing retrieves more than one document containing relevant information for the answer, even if not spot on. This is because many questions in our newly developed test set are open-ended and opinions about the correct answer vary. There were questions with contradictory answers that were being marked as correct since it is not the duty of (this) QA system to infer the correct answer to a input question, but merely to present likely options as possible answers. One such example is the question ``Ajută vitamina D la prevenirea COVID?'' (``Does vitamin D help preventing COVID infections?'') for which we find that either ``Lipsa vitaminei D crește riscul de infecție, inclusiv cu SARS-Cov-2.'' (``The lack of vitamin D causes a rise in the risk of a SARS-Cov-2 infection.'') or ``... dovezile existente nu susțin eficacitatea vitaminei D pentru tratarea virusului Covid-19.'' (``... existing evidence does not support the efficiency of vitamin D in treating COVID-19'').

Consequently, our 65-question test set contains, for each question, a list of the URLs of the documents containing the relevant and recent (if necessary) answer. For each URL, we retain the Bing-extracted text snippet in which we manually highlight one or more likely answers. For 10 questions out of 65, our automatic query generation algorithm did not find any suitable documents. We can thus evaluate the effectiveness of the content word selection query generation algorithm at $1 - \frac{10}{65} \approx 85\%$.

Table \ref{tab:qa_test_results} presents the results of the QA system, on the test set presented above, using the baseline query generation algorithm (the input question itself) vs. the content word (CW) selection query generation algorithm. We compute the following:

\begin{itemize}
    \item \textbf{Mean Reciprocal Rank} (MRR) of the returned documents: if multiple documents contain relevant information, we choose the one with the highest ranking to contribute to the MRR. The text snippets retrieved by Bing are re-ranked using Equation \ref{eq:3}, which provides the order in which the user sees the results.
    \item \textbf{Exact answer matching}: percentage of BERT highlighted answers that exactly match a human highlighted answer in the test set. Because our QA system only highlights a single answer in the returned text snippet, if the question has multiple answers that are annotated as correct, we test each annotated answer for an exact match.
    \item \textbf{F1 overlap matching}: if an exact match does not exist between the BERT answer and any of the annotated answers, we find the annotated answer that has the highest overlap with the BERT answer and compute the F1 measure of the overlapped characters.
\end{itemize}

\begin{table}[]
    \centering
    \begin{tabular}{lccc}
        \hline
        \textbf{Query gen.} & \textbf{MRR} & \textbf{Exact \%} & \textbf{F1 \%} \\
        \hline
        Baseline query gen. & 0.4056 & 33.85 & 65.07 \\
        CW query gen. & 0.5337 & 50.77 & 76.69 \\
        \hline
    \end{tabular}
    \caption{Results of the QA system on the new test set}
    \label{tab:qa_test_results}
\end{table}

Table \ref{tab:qa_test_results} shows convincingly that the content word selection query generation algorithm is much better than using the input question as the Bing query. A MRR of more than $0.5$ shows that the user sees the snippet containing a likely answer in the top two results returned by the QA system. It is also encouraging that the F1 overlap score for the BERT answer highlight algorithm is on par with the F1 obtained when training on the data set presented in Section \ref{sec:dataset} (see Table \ref{tab:qa_results}), even if the newly developed test set contains more recent questions than the ones in that data set.

\section{Conclusions}
\label{sec:conclusions}

We have presented an open-domain QA system that uses a fine-tuned BERT model to highlight probable answers to the input question in the Bing-returned text snippets. With a MRR bigger than $0.5$, we have the guarantee that the user sees relevant content in the first two results returned by the QA system, with more relevant content following, as Bing finds useful information in more than one document. Furthermore, a character overlap F1 of almost 77\% between the correct answer and the BERT supplied answer will steer user's attention effectively towards the correct answer.

Comparing the data sets sizes, we see that our COVID-19 data set is two orders of magnitude smaller than the data sets presented in \citet{rajpurkar2018squad}. The exact match and F1 scores on the SQuAD 1.1 data set are 78.6\% and 85.8\% respectively, suggesting that we have room to improve in these areas, provided that we can grow our data set significantly. But even with the current performance, the QA system is useful as it is.

The open-domain QA pipeline is trainable in the sense that, given a data set similar to the one presented in Section \ref{sec:dataset} but in a different domain, one can fine-tune the chosen BERT model to answer questions in the domain of the new data set. Relying on a web search engine such as Bing, indexing billions of documents on a regular basis, the part of answer retrieving is assured, irrespective of the chosen domain. 

In a different use case, the QA system can be adapted to work on e.g. public institution web sites, by having the public institution web sites indexed using either a local search engine or local Bing/Google indexing and using these specialized search results instead of the web-wide search results. Coupled with a data set of questions on the specific topic of interest (e.g. ``tax payments'', ``public transportation'', ``resident parking'', etc.), the QA pipeline can work to answer targeted questions from citizens.

The next steps in the development of this QA system are:

\begin{itemize}
    \item To test it with other languages by using the \emph{eTranslation} online machine translation service provided by the European Commission. We could automatically translate the input question into Romanian, run the QA pipeline and then translate the output of the QA system into the input question language.
    \item To develop an answer mining algorithm that is better than Bing's algorithm for mining the text snippet that is most relevant to the query. We would parse the URL of the returned document and select the text snippet that contains the relevant answer ourselves, instead of using the provided text snippet.
\end{itemize}

\section*{Acknowledgments}
The Action 2020-EU-IA-0088 (``Enrich4All'') has received funding from the European Union's Connecting Europe Facility 2014-2020 -- CEF Telecom, under Grant Agreement No. INEA/CEF/ICT/A2020/2278547.

\bibliographystyle{clib_acl_natbib}
\bibliography{clib}

\end{document}